\pdfoutput=1
\documentclass[11pt]{article}

\usepackage{acl}

\usepackage{times}
\usepackage{latexsym}

\usepackage{graphicx}

\usepackage[T1]{fontenc}
\usepackage[utf8]{inputenc}

\usepackage{microtype}

\usepackage{booktabs}

\usepackage{amsmath,amsthm,amssymb,amsfonts,mathtools,verbatim,xcolor}

\newcommand{\R}{\ensuremath{\mathbb R}}
\newcommand{\T}{\top}
\newcommand{\DMU}{\textrm{DMU}}

\title{Assessing Resource-Performance Trade-off of Natural Language Models using Data Envelopment Analysis}%{Data Envelopment Analysis for Natural Language Processing}

\author{Zachary Zhou \\
Industrial and Systems Engineering\\
University of Wisconsin -- Madison\\
1415 Engineering Drive\\
Madison, WI 53706\\
{\tt zzhou246@wisc.edu}
\And
    Alisha Zachariah \\
    {\bf Devin Conathan} \\
    {\bf Jeffery Kline} \\
    American Family Insurance \\
    6000 American Parkway \\
    Madison, WI 53783\\
    {\tt \{alisha044,dconathan,jeffery.kline\}@gmail.com}
    }

\begin{document}
\maketitle
\begin{abstract}

Natural language models are often summarized through a high-dimensional set of descriptive metrics including training corpus size, training time, the number of trainable parameters, inference times, and evaluation statistics that assess performance across tasks. The high dimensional nature of these metrics yields challenges with regard to objectively comparing models; in particular it is challenging to assess the trade-off models make between performance and resources (compute time, memory, {etc.}).

We apply Data Envelopment Analysis (DEA) to this problem of assessing the resource-performance trade-off. DEA is a nonparametric method that measures productive efficiency of abstract {\em units} that consume one or more {\em inputs} and yield at least one {\em output}.  We recast natural language models as units suitable for DEA, and we show that DEA can be used to create an effective framework for quantifying model performance and efficiency.  A central feature of DEA is that it identifies a subset of models that live on an {\em efficient frontier} of performance. DEA is also scalable, having been applied to problems with thousands of units. We report empirical results of DEA applied to 14 different language models that have a variety of architectures, and we show that DEA can be used to identify a subset of models that effectively balance resource demands against performance.
\end{abstract}

\section{Introduction}

A standard task in the machine learning lifecycle is to compare performance of many models; typically this process involves analyzing high-dimensional sets of summary statistics (hyperparameters, evaluation metrics, {etc.}). A common use case is quantifying the trade-off between performance and resource constraints; the goal being to achieve the best possible performance using minimal resources.

%If each model in the collection is summarized using a single real-valued scalar, then one can simply rank-order all models according to that scalar's value.  But more common, models are described through a a dimensional set of metrics.

Meanwhile, multitask performance benchmarks ({e.g.,}\ GLUE) have found widespread adoption in the natural language processing (NLP) community, with transformer-based models often leading in evaluation performance \cite{NIPS2017_3f5ee243}. While the performance of transformer-based  language models is impressive, they are notoriously resource-intensive, and often smaller models can more efficiently leverage a limited resource budget.  However, it is nontrivial to demonstrate this fact by formulating a rational and fair comparison among models of different sizes and architectures.

\begin{figure*}[!ht]
    \centering
    \includegraphics[width=0.95\textwidth]{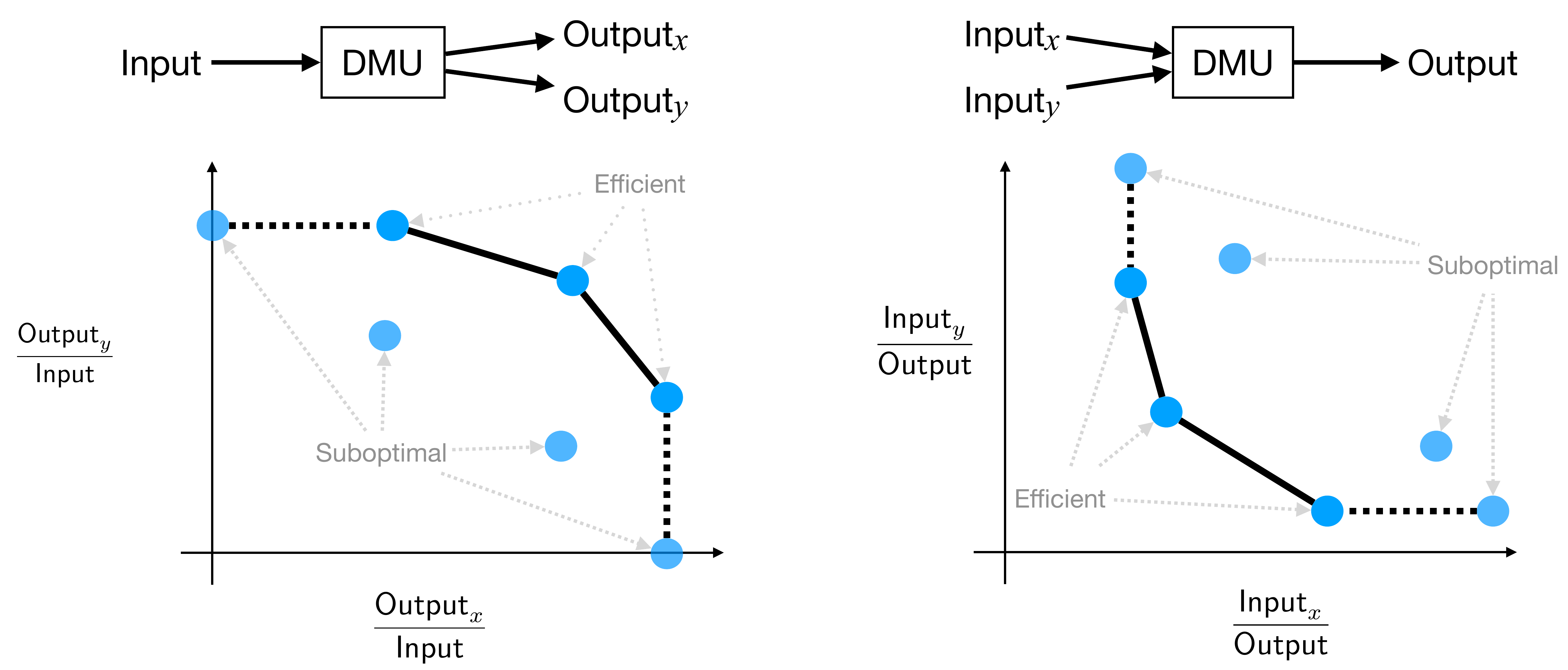}
    \caption{Two simple examples of DEA. ({\em{Left}}) Each DMU, represented by a blue dot, has two outputs and a single input. Efficient DMUs generally lie far from the origin, which corresponds to DMUs that have high output per unit of input.   ({\em{Right})} Each DMU ingests two inputs and yields a single output. The set of DMUs that lie close to either axis are more efficient, which has the interpretation of low unit input per unit of output.  The Pareto fronts are indicated with the heavy black line segments. Note that the dashed lines segments are not considered part of the front. Suboptimal DMUs are said to be ``enveloped'' by efficient DMUs.}
    \label{fig:fig2and1}
\end{figure*}

% The set of models may be the result of a hyperparameter search, or else simply from a survey of competing architectures. Since state-of-the-art models are highly complex objects, effective summary descriptions include many features that describe aspects of training, task performance and resource consumption.

% But the problem faced when trying to select the optimal model from among many competing notions of either success or cost, and one often has several competing metrics that assess performance. 

% But performance metrics are themselves imperfect, and a common solution to this is to introduce a new performance metric, which adds new metrics to a crowded field.  

% But when assessing these trade-offs, 
% It is very common to see model metrics reported in large tables, and in such cases, the trade-off space is navigated using heuristics and ad-hoc inspection. 

% reported in diverse units ({\em e.g.}, seconds, bytes, accuracy), and assembling them into a unified, normalized structure 

% and models are compared using data that is represented a large, unordered set of options, where the attributes of the models vary by loss, size, training time, and performance metrics. 
In this paper, we apply {\em data envelopment analysis} (DEA) to this challenge of assessing model resource-performance trade-off \cite{charnes1978measuring,Banker1984}. DEA is a technique that originated in the operations research community, and it has been applied to a wide range of settings over many decades.  It is traditionally concerned with rigorously defining {\em decision making efficiency} for teams, departments, companies, and other types of people-oriented organizations.   But DEA is a {generic} technique that is based on solving a series of linear programs that are constructed to analyze the relative efficiency of {\em decision making units} (DMUs). A DMU is an abstract object that converts a set of {\em inputs} or {\em resources} into a set of {\em outputs} or {\em benefits}. %DEA is an analysis that is applied to the inputs and outputs of a population of DMUs, and it assigns a scalar value between 0 and 1, representing efficiency, to each DMU. A DMU that is more effective at transforming inputs into outputs is considered more efficient.

Our adaptation of DEA to the NLP context begins by treating each model as a DMU. %, and we identify the sets of inputs and outputs associated with each model with the metrics that describe said model. The analysis ends with a report of each model's efficiency in transforming the model inputs into the model's outputs.
Example inputs for the analysis include training time, training corpus size, the number of trainable parameters, and total monetary cost to train. Typical outputs would be performance evaluation metrics, evaluation throughput, {etc}. %In general, an increase in an output's magnitude improves the overall value or utility of the model. A model {\em input} is a measurement characterized by the property that an increase in its magnitude should causally increase the value of at least one model output. Additionally, each input may  serve as a proxy for consuming a resource such as time, money or space.

Our main contribution in this paper is that we apply DEA to the problem of assessing the resource-performance trade-off of machine learning models with an emphasis on evaluating the efficiency of  language models. To our knowledge, this application of DEA to machine learning has not appeared in prior work. We do not assume familiarity with DEA, so in Section~\ref{sec:math-setup} we provide sufficient detail to interpret our empirical results, which apply DEA to a variety of models, in Section~\ref{sec:results}.

\section{Background}
DEA was developed to enable performance assessment of teams of people and organizations such as not-for-profits, governmental organizations, departments within larger organizations and meta-analyses of industries. Traditional inputs include organizational staff salary, operational costs, and time. Traditional outputs include revenue, sales volume, and other organizational goals. The Pareto front of DEA in this context is also known as the {\em best practice frontier}, with the name derived from the observation that if a decision making unit (DMU) is on this frontier, it is objectively more efficient at transforming its inputs into outputs.

DEA analysis is applied to the inputs and outputs of a population of DMUs, and it assigns a scalar value between 0 and 1 to each DMU which expresses its efficiency.  A DMU that is more effective at transforming inputs into outputs is considered more efficient.

Figure~\ref{fig:fig2and1} illustrates two  simple scenarios where DEA-type efficiency can be endowed with an intuitive representation.  On the left, several DMUs are represented as blue dots, and each DMU ingests a single input and yields two outputs.  A process that generates more output for a given input is considered to have greater efficiency. In this scenario, processes that lie on the Pareto front are far from the origin.  At right, a different set of DMUs is shown. In this example, each DMU ingests two inputs and performance is assessed through a single output. A DMU with low input or large output will live  close to one of the axes, and DMUs close to the origin are more efficient.

We now illustrate how DEA quantifies model efficiency by briefly describing a hypothetical, and simple, example. Consider a language model trained on a small amount of data with high accuracy for some task. DEA classifies this model as {\em more efficient} than (1) a model that achieves the same accuracy with more training data or (2) a model that achieves a lower accuracy with the same amount of training data.

% DEA generalizes to apply to units that have many inputs and many outputs, though visualization as shown in Figure~\ref{fig:fig2and1} in such cases is not usually possible. The result of performing DEA on a set of models is to assign each model an efficiency score, which is a scalar value between 0 and 1, as well as coefficients that express the performance of each model in terms of the Pareto-optimal models. Through DEA, one converts a multidimensional description of  many models into a rank-ordered list of models via the efficiency metric, and the performance of all other models can be expressed in terms of the Pareto-optimal set.

We describe the formal definition of DEA below in Section~\ref{sec:math-setup}, but for now it suffices to understand that DEA is the result of solving a sequence of linear programs.  In particular, global solutions are guaranteed to be found rapidly and with high numerical precision.

% In general, an increase in an output's magnitude improves the overall value or utility of the model. A model {\em input} is a measurement characterized by the property that an increase in its magnitude should causally increase the value of at least one model output. Additionally, each input may  serve as a proxy for consuming a resource such as time, money or space.

A DEA-based approach to model comparison has several advantages. Since it is based on linear programming, the DEA framework lends itself to detailed theoretical analysis, which extends to interpreting solutions and modifying the programs in a controlled way.  Furthermore, DEA is extensible both in the number of models that one can consider as well as the metrics that are used to represent each model's inputs and outputs.  Finally, DEA is a scalable technique, since it allows one to analyze model performance of tens of thousands of models.

% Increasing performance of models, larger capabilities and training sets, making a principled decision is challenging. 

% In machine learning scenarios, one must decide which model to use, balancing judgements about sometimes competing model performance metrics. The operations research community has formulated a principled framework for analyzing this scenario, and work extends back several decades  \cite{keeney1993decisions}.  

% An advantage of machine learning applications over the other domains where the field emerged is the very quantitative nature of metrics: often the metrics are expressed in absolute terms, and all attributes can be precisely measured, including uncertainties.  Thus, subjectivities can be eliminated. 

% However, numerical precision does not imply that the optimal choice is necessarily objective. 

 DEA can be applied to almost arbitrary data that satisfies a small number of weak conditions, but this flexibility comes with some cost. In order to derive meaning from DEA, one must carefully choose the set of inputs and outputs. This process of selection is necessarily subjective. The specifics of our implementation are not meant to be universal prescriptions but rather a demonstration of the concept and useful starting point.

\section{Related work}
% Links to survey  GLUE
% F1 score references
% [Chinchor, 1992] Nancy Chinchor, MUC-4 Evaluation Metrics, in Proc. of the Fourth Message Understanding Conference, pp. 22–29, 1992. http://www.aclweb.org/anthology-new/M/M92/M92-1002.pdf
% [van Rijsbergen, 1979] C. J. van Rijsbergen, Information Retrieval, London: Butterworths, 1979. http://www.dcs.gla.ac.uk/Keith/Preface.html
% \begin{itemize}
%     \item the history of DEA (CCR and BCC)
%     \item applications of DEA in operations research
%     \item examples of heterogeneous model performance metrics
%     \item mention the F1 score?
% \end{itemize}

\label{sec:related}
 
DEA was introduced to the operations research community as a tool to help organizations quantify efficiency, and to objectively identify  suborganizations that perform especially well~\cite{charnes1978measuring}.  
Since its introduction, DEA has been applied to a vast array of fields, including international banking, cloud computing operations, economic sustainability, police department operations, hospital operations, and logistical applications~\cite{charnes1995data,dea2016aliraylei, SUN200251,THANASSOULIS1995641,dea4ml2}. Recent work has applied DEA to the machine learning context to optimize generalization error of models~\cite{dea4ml}; we are unaware of prior work that applies DEA to the purpose of assessing model efficiency as we do here.

The theory of DEA continues to be an active field of research, and there have been many developments over the years in an attempt to address perceived shortcomings.   In addition to the relaxation of constant returns to scale, ``cross-efficiency'' was introduced to generate unique efficiency rankings, and ``stochastic data envelopment analysis'' was developed to account for noise and uncertainty in the measurements  that are used to inform DEA~\cite{Banker1984,doyle1994efficiency,OLESEN20162}.

DEA is parallelizable, and it has been applied to problems with tens of thousands of DMUs~\cite{PHILLIPS1990167,KHEZRIMOTLAGH20191047}. Reducing the required computation time of DEA has also been explored~\cite{ALI1990157,ALI199361}.

Assessment of natural language understanding requires models to execute a range of linguistic tasks across different domains.  
Recognizing this, the {GLUE} benchmarks were introduced~\cite{wang-etal-2018-glue}.  The GLUE benchmarks consist of nine English sentence understanding tasks, so the performance of a single model on the GLUE benchmark yields a nine-dimensional vector. Typically, this vector is summarized through an average and reported as a single score.

A challenge of modern transformer-based machine learning is the large number of different architectures that can be tested. A fairly comprehensive  overview of recent performance results related to language models is reported in \cite{narang2021transformer}. The primary method of reporting the results is in tabular form (see, for example Tables 1 and 2 of that reference), and comparative analysis is challenging.  Others propose rigorous scientific methods and experiment design to help manage these challenges
\cite{ulmer-experimental-standards}, \cite{dror-etal-2019-deep}, \cite{dror-replicability-analysis}; we believe DEA is another tool that can be leveraged for these analyses.

Multidimensional descriptions of models are an inescapable feature of machine learning, and scalarization of such descriptions are equally common. Several metrics commonly used to describe models include precision, recall, accuracy, model size, and a variety measures of performance including BLEU and the family of ROUGE scores~\cite{rouge,papineni2002bleu}. Aside from DEA, other well-known examples of scalarization techniques used within the machine learning community include the F1 score, the Matthews correlation coefficient, and the Fowlkes–Mallows index, which summarizes the confusion matrix~\cite{MATTHEWS1975442,yule,fowlkesmallows}. % Recent extensions of these foundational ideas appeared~\cite{Yacouby2020ProbabilisticEO}.

\section{Mathematical background}
\label{sec:math-setup}
% DEA was developed to enable performance assessment of teams of people and organizations such as not-for-profits, governmental organizations, departments within larger organizations and meta-analyses of industries. Traditional inputs include organizational staff salary, operational costs, and time. Traditional outputs include revenue, sales volume, and other organizational goals. The Pareto front of DEA in this context is also known as the {\em best practice frontier}, with the name derived from the observation that if a DMU is on this frontier, it is objectively more efficient at transforming its inputs into outputs.  

%DEA is used to identify DMUs that are highly effective at transforming inputs into outputs, and to discover the optimal DMUs and transfer aspects of their operational activities to other suboptimal DMUs. 

%DEA formalizes the simple idea that for an arbitrary process, more input should yield more output. 
In this section, we provide sufficient background for one who is unfamiliar with DEA to interpret the  results of Section~\ref{sec:results}.
 
When DEA was originally introduced, a technical requirement of the processes being assessed was that they exhibit {\em  constant returns to scale}; for example this means that doubling the value of each input (e.g., sales staff) should cause the doubling of the value of all the outputs (e.g., monthly sales). DEA found widespread adoption despite this assumption almost never holding in practice. To address this perceived deficiency in DEA, the original formulation of DEA was modified to relax the constant returns to scale assumption.  Several other extensions of DEA are now in common use, and we provide an overview below. %here. 
More %complete 
details can be found in~\cite{cooper2007data}. 

We introduce the setup and notation as follows. There are $n$ DMUs, each of which consumes $m$ inputs and produces $s$ outputs.
Concretely, $\DMU_j$ consumes $x_{ij} \ge 0$ units of input and produces $y_{rj} \ge 0$ units of output, where $1\leq i\leq m$, $1\leq r\leq s$, and $1\leq j \leq n$. The measurement units of the different inputs and outputs need not be congruent.
For shorthand, we can express the data corresponding to $\DMU_j$ with the pair $(x_j,y_j)\in\R^{m + s}$ where $x_j = (x_{ij})_{i=1}^m$ and $y_j = (y_{rj})_{r=1}^s$. We call the pair $(x_j, y_j)$ an {\em activity}.
We can additionally arrange the input data in an $m\times n$ matrix $X = (x_{ij})$ and the output data in an $s\times n$ matrix $Y = (y_{rj})$.
All the vectors $x_j$ and $y_j$ are assumed to be \emph{semipositive}, meaning their entries are nonnegative and at least one entry is strictly positive. Equivalently, this means $\DMU_j$ consumes a positive amount of some input and produces a positive amount of some output.

\subsection{CCR efficiency}
We first introduce the (input-oriented) CCR model \cite{charnes1978measuring}, named as such after its creators Charnes, Cooper, and Rhodes. The CCR model is widely regarded as the first DEA model, and assumes constant returns to scale. 

For each $o$, where $1\leq o \leq n$, we evaluate  $\DMU_o$~against its peers. Let $v = (v_i)_{i=1}^m \in \R_+^m$ and $u = (u_r)_{r=1}^s \in \R_+^s$ denote the weights that are applied to all the inputs and all the outputs of $DMU_o$, respectively.
For an arbitrary activity $(x,y) \in \R_+^{m+s}$, the ratio $u^\T y / v^\T x$ measures efficiency by reducing the multiple inputs (resp. outputs) to a single ``virtual'' input (resp. ``virtual'' output), then returning the ratio of virtual output to virtual input.
The CCR model aims to solve the following fractional program, indexed by $o$, where $1\leq o\leq n$:
\begin{subequations} \label{CCR fractional}
\begin{align}
\underset{v,u}{\text{maximize}} \quad
& \theta\coloneqq \theta_o = \frac{u^\T y_o}{v^\T x_o} \label{CCR fractional objective} \\
\text{subject to} \quad
& \frac{u^\T y_j}{v^\T x_j} \le 1 \quad \text{ for } j=1,\dots,n \label{CCR fractional constraints} \\
& v \in \R_+^m,\ u \in \R_+^s.
\end{align}
\end{subequations}
The constraints \eqref{CCR fractional constraints} bound the efficiency ratio of each DMU above by 1.
The objective \eqref{CCR fractional objective} aims to find multipliers $v,u$ that maximize the efficiency ratio of target $\DMU_o$; due to the constraints \eqref{CCR fractional constraints}, clearly the optimal value $\theta^*$ is at most 1.
It can be shown that Eq.~\eqref{CCR fractional} is equivalent to the following linear program,  called the CCR multiplier form:
\begin{subequations} \label{CCR multiplier}
\begin{align}
\underset{v,u}{\text{maximize}} \quad
& \theta = u^\T y_o \\
\text{subject to} \quad
& v^\T x_o = 1 \\
& -v^\T X + u^\T Y \le 0^\T \\
& v \in \R_+^m,\ u \in \R_+^s.
\end{align}
\end{subequations}
Equivalence of \eqref{CCR fractional} and \eqref{CCR multiplier} can be verified through a simple exercise \cite{cooper2007data}.
We call $\DMU_o$  \emph{CCR-efficient} if $\theta^*=1$ and there exists an optimal $(v^*,u^*)$ with $v^* > 0$ and $u^* > 0$. Otherwise we call $\DMU_o$ \emph{CCR-inefficient}.

It is possible for $\DMU_o$ to achieve the maximal value $\theta^*=1$ and still be CCR-inefficient; this occurs when some $\DMU_j \ne \DMU_o$ consumes no more input than $\DMU_o$, produces at least as much output as $\DMU_o$, and either consumes strictly less of some input or produces strictly more of some output than $\DMU_o$.
In the literature, such CCR-inefficient DMUs are occasionally referred to as \emph{weakly efficient}, whereas DMUs satisfying both $\theta^*=1$ and $(v^*,u^*) > 0$ are called \emph{strongly efficient} \cite{handbook2004}.
In Figure~\ref{fig:fig2and1}, the weakly inefficient points are the endpoints of the dashed line segments that are parallel to the axes, and they are labeled ``suboptimal.''
For the most part, we will not use this terminology and simply refer to DMUs satisfying $\theta^*=1$ and not $(v^*,u^*) > 0$ as inefficient.

Computationally, one typically does not work with the CCR multiplier form directly, but rather with its dual.
The dual of \eqref{CCR multiplier} is referred to as the {\em CCR envelopment form}:
\begin{subequations} \label{CCR envelopment}
\begin{align}
\underset{\theta,\lambda}{\text{minimize}} \quad
& \theta \\
\text{subject to} \quad
& \theta x_o - X\lambda \ge 0 \label{CCR input constr} \\
& Y\lambda \ge y_o \label{CCR output constr} \\
& \theta \in \R,\ \lambda \in \R_+^n.
\end{align}
\end{subequations}
% FIXME: Possibly move BCC (5) closer to CCR (3), move production possibility sets to appendix

We now describe the connection between the CCR model and the assumption of constant returns to scale with an alternative interpretation of the envelopment form.
Recall that an arbitrary pair of vectors $(x,y) \in \R_+^{m+s}$ is called an \emph{activity}.
The CCR model assumes there is a set of feasible activities, called the \emph{production possibility set}, denoted $P_{CCR}$, which is defined as the polytope
\begin{multline*}
P_{CCR} \coloneqq \{(x,y) \in \R_+^{m+s} : \\ x \ge X\lambda,\ y \le Y\lambda,\ \lambda \in \R_+^n\},
\end{multline*}
and which has the following properties:
\begin{enumerate}
    \item We assume the observed activities $\{(x_j,y_j)\}_{j=1}^n$ are contained in $P_{CCR}$.
    \item If $(x,y) \in P_{CCR}$, then $(\bar x,\bar y) \in P_{CCR}$ for any $\bar x \ge x,\ \bar y \ge y$. (In the economics literature, this is known as \emph{free disposability} \cite{Carter_1952}.)
    \item Conic combinations of activities in $P_{CCR}$ belong to $P_{CCR}$.
\end{enumerate}
The last property implies constant returns to scale, as $(x,y) \in P_{CCR}$ implies $(tx,ty) \in P_{CCR}$ for any $t > 0$.

Eq.~\eqref{CCR envelopment} can be viewed as finding the minimum $\theta$ such that $(\theta x_o,y_o) \in P_{CCR}$.
More intuitively, Eq.~\eqref{CCR envelopment} aims to synthesize a new activity using conic combinations of the observed activities $\{(x_j,y_j)\}_{j=1}^n$, i.e., $(X\lambda,Y\lambda)$ where $\lambda \in \R_+^n$.
Eq.~\eqref{CCR envelopment} tries to scale the inputs $x_o$ as small as possible by the factor $\theta$ while ensuring that the synthesized activity $(X\lambda,Y\lambda)$ consumes no more inputs than $\theta x_o$ and maintains output levels at least as high as $y_o$.

The envelopment form allows for an alternative characterization of CCR-efficiency: $\DMU_o$ is CCR-efficient if for any optimal solution $(\theta^*,\lambda^*)$ to \eqref{CCR envelopment}, $\theta^*=1$ and the solution has zero slack, i.e., the constraints \eqref{CCR input constr} and \eqref{CCR output constr} hold at equality; $\DMU_o$ is CCR-inefficient otherwise.
%This definition of CCR-efficiency is equivalent to the former definition involving the multiplier form \eqref{CCR multiplier}.
If $\DMU_o$ is CCR-inefficient, then there exists $\lambda \in \R_+^n$ such that $x_o \ge X\lambda,\ Y\lambda \ge y_o$ and at least one inequality in the system holds strictly; the synthesized activity $(X\lambda,Y\lambda)$ is thus strictly better than $(x_o,y_o)$, and so $\DMU_o$ is said to be \emph{enveloped} by the observed activities $\{(x_j,y_j)\}_{j=1}^n$.

Solving \eqref{CCR envelopment} alone is not enough to determine whether $\DMU_o$ is CCR-efficient; to determine whether every optimal solution to \eqref{CCR envelopment} has zero slack, one additionally solves the following linear program:
\begin{subequations} \label{CCR envelopment slacks}
\begin{align}
\underset{\lambda,s^-,s^+}{\text{maximize}} \quad
& 1^\T s^- + 1^\T s^+ \\
\text{subject to} \quad
& s^- = \theta^*x_o - X\lambda \\
& s^+ = Y\lambda - y_o \\
& \lambda \in \R_+^n,\ s^- \in \R_+^m,\ s^+ \in \R_+^s,
\end{align}
\end{subequations}
where $\theta^*$ in \eqref{CCR envelopment slacks} is the optimal value of \eqref{CCR envelopment}.
If $\DMU_o$ is CCR-inefficient, we can additionally find its \emph{reference set}, the set of CCR-efficient DMUs that envelop $\DMU_o$ thus making it CCR-inefficient. The reference set is defined based on the max-slack solution $(\theta^*,\lambda^*,s^{-*},s^{+*})$ of \eqref{CCR envelopment} and \eqref{CCR envelopment slacks} to be \[E^{CCR}_o = \{j \in \{1,\dots,n\} : \lambda^*_j > 0\}.\]

\subsection{BCC efficiency}
The constant returns to scale assumption of the CCR model can be problematic when comparing language models, e.g., one typically expects diminishing returns from increased training time.  Fortunately, this can be relaxed with a very simple modification to the CCR formulation~\cite{Banker1984}.
The so-called BCC model, named after its creators Banker, Charnes, and Cooper, addresses this shortcoming and allows for variable returns to scale by adding a single additional constraint, namely $1^\T\lambda = 1$, on the production possibility set.
The BCC envelopment form, which is almost identical to Eq.~\eqref{CCR envelopment}, is as follows:
\begin{subequations} \label{BCC envelopment}
\begin{align}
\underset{\theta,\lambda}{\text{minimize}} \quad
& \theta \\
\text{subject to} \quad
& \theta x_o - X\lambda \ge 0 \label{BCC input constr} \\
& Y\lambda \ge y_o \label{BCC output constr} \\
& 1^\T\lambda = 1 \\
& \theta \in \R,\ \lambda \in \R_+^n.
\end{align}
\end{subequations}
 The dual of \eqref{BCC envelopment} is the BCC multiplier form:
\begin{subequations} \label{BCC multiplier}
\begin{align}
\underset{v,u,u_0}{\text{maximize}} \quad
& u^\T y_o - u_0 \\
\text{subject to} \quad
& v^\T x_o = 1 \\
& -v^\T X + u^\T Y - u_0 1^\T \le 0^\T \\
& v \in \R_+^m,\ u \in \R_+^s,\ u_0 \in \R.
\end{align}
\end{subequations}
The production possibility set $P_{BCC}$ of the BCC model is defined as
\begin{multline*}
P_{BCC} = \{(x,y) \in \R_+^{m+s} : \\ x \ge X\lambda,\ y \le Y\lambda,\ 1^\T\lambda = 1,\ \lambda \in \R_+^n\}.
\end{multline*}
The envelopment form \eqref{BCC envelopment} can be viewed as finding the minimum $\theta$ such that $(\theta x_o,y_o) \in P_{BCC}$.
We call $DMU_o$ \emph{BCC-efficient} if for any optimal solution $(\theta^*,\lambda^*)$ to \eqref{BCC envelopment}, $\theta^*=1$ and the solution has zero slack, i.e., the constraints \eqref{BCC input constr} and \eqref{BCC output constr} hold at equality; $\DMU_o$ is \emph{BCC-inefficient} otherwise.
As in the case of the CCR model, one not only solves \eqref{BCC envelopment} but also the following:
\begin{subequations} \label{BCC envelopment slacks}
\begin{align}
\underset{\lambda,s^-,s^+}{\text{maximize}} \quad
& 1^\T s^- + 1^\T s^+ \\
\text{subject to} \quad
& s^- = \theta^*x_o - X\lambda \\
& s^+ = Y\lambda - y_o \\
& 1^\T\lambda = 1 \\
& \lambda \in \R_+^n,\ s^- \in \R_+^m,\ s^+ \in \R_+^s,
\end{align}
\end{subequations}
where $\theta^*$ in \eqref{BCC envelopment slacks} is the optimal value of \eqref{BCC envelopment}.

If $\DMU_o$ is BCC-inefficient, we are interested in finding its \emph{reference set}, the set of BCC-efficient DMUs that envelop $\DMU_o$ thus making it BCC-inefficient. The reference set is defined based on the max-slack solution $(\theta^*,\lambda^*,s^{-*},s^{+*})$ of \eqref{BCC envelopment} and \eqref{BCC envelopment slacks} to be \[E^{BCC}_o = \{j \in \{1,\dots,n\} : \lambda^*_j > 0\}.\]
\label{sec:background}
If $\DMU_o$ is BCC-efficient, we can additionally determine returns to scale as follows:
\begin{enumerate}
    \item Increasing returns to scale prevails at $(x_o,y_o)$ iff $u_0^* < 0$ \emph{for all} optimal solutions to \eqref{BCC multiplier}.
    \item Decreasing returns to scale prevails at $(x_o,y_o)$ iff $u_0^* > 0$ \emph{for all} optimal solutions to \eqref{BCC multiplier}.\label{item:drs}
    \item Constant returns to scale prevails at $(x_o,y_o)$ iff $u_0^* = 0$ \emph{for some} optimal solution to \eqref{BCC multiplier}.\label{item:crs}
\end{enumerate}
Suppose we solve \eqref{BCC multiplier} and obtain $u_0^* < 0$. We then solve the following modified program:
\begin{subequations} \label{BCC returns to scale}
\begin{align}
\underset{v,u,u_0}{\text{maximize}} \quad
& u_0 \\
\text{subject to} \quad
& v^\T x_o = 1 \\
& u^\T y_o - u_0 = 1 \\
& -v^\T X + u^\T Y - u_0 1^\T \le 0^\T \\
& v \in \R_+^m,\ u \in \R_+^s,\ u_0 \le 0.
\end{align}
\end{subequations}
If \eqref{BCC returns to scale} yields an optimal solution with $u_0^* = 0$, then constant returns to scale prevails at $(x_o,y_o)$, otherwise increasing returns to scale prevails.
If on the other hand we solve \eqref{BCC multiplier} and obtain $u_0^* > 0$, \eqref{BCC returns to scale} can be modified by replacing the constraint $u_0 \le 0$ with $u_0 \ge 0$ and switching the optimization sense to minimize $u_0$.

Since the BCC envelopment form differs from the CCR envelopment form only in the addition of the convexity constraint $1^\T\lambda = 1$, if $\DMU_o$ is CCR-efficient, it is also BCC-efficient, and constant returns to scale prevail at $\DMU_o$.

%The CCR and BCC scores of a DMU can be used to understand the sources of inefficiency.
The CCR score $\theta_{CCR}^*$ is called the \emph{(global) technical efficiency ($TE$)} as the CCR model ignores the effects of scaling.
The BCC score $\theta_{BCC}^*$ is called the \emph{(local) pure technical efficiency ($PTE$)} as the BCC model accounts for variable returns to scale.
The \emph{scale efficiency ($SE$)} is defined as 
\begin{align}SE = \frac{TE}{PTE} = \frac{\theta_{CCR}^*}{\theta_{BCC}^*}.
\label{Scale efficiency}
\end{align}
Note that $0 \le SE \le 1$.
Eq.~\eqref{Scale efficiency} implies a decomposition of technical efficiency into pure technical efficiency and scale efficiency; if technical efficiency $TE$ is low, it is either because of inefficient operation (low $PTE$) or poor scaling of resources (low $SE$).

We remark that all of the CCR and BCC models we consider are \emph{input-oriented}, as they attempt to reduce input consumption while maintaining the same if not higher level of output production.
We do not consider \emph{output-oriented} models which consider the opposite situation where output production is increased while maintaining the same or lower level of input consumption.

\section{Results and analysis}
\label{sec:results}
In this section, we describe the results of applying DEA to compare a variety of NLP models. The input features and the output features were selected to incorporate aspects of training, evaluation and task performance. Since training is one part of our analysis, several identical versions of the same models are  represented in the set of models considered but with different learning rates selected. We also incorporate several simpler models as baselines including TF-IDF and GloVe embeddings with linear classifiers~\cite{pennington2014glove}.

The transformer models are pretrained~models that are sourced from the Hugging Face Model Hub~\cite{huggingface}.
%\footnote{\url{https://huggingface.co/models}}. 
Each transformer model appears three times: once for each of the learning rates $10^{-3}$, $10^{-4}$, and $10^{-5}$. The~base models are \texttt{bert-base-uncased},
\texttt{bert-large-uncased} \cite{DBLP:conf/naacl/DevlinCLT19}, \texttt{roberta-base} \cite{DBLP:journals/corr/abs-1907-11692}, and their distilled versions: \texttt{distilbert-base-uncased}, \texttt{distilroberta-base} \cite{DBLP:journals/corr/abs-1910-01108}.
The GloVe embeddings used are all trained on the Wikipedia 2014 and Gigaword 5 6B corpuses and vary in embedding dimension between 50, 100, 200 and 300 \cite{pennington2014glove}.
%\footnote{\url{https://nlp.stanford.edu/projects/glove/}}.
As our simplest baseline we use \texttt{scikit-learn}'s implementation of TF-IDF which varies in vocabulary size between 100, 500,  1000, 5000, 10000 and 15000 \cite{scikit-learn}. 

The number of trainable parameters for the transformer and other deep network models is determined by the model architecture and is typically in the millions.  For the simpler embedding-based models, the number of trainable parameters is determined by the embedding dimension or vocabulary size. The GLUE benchmarks were coalesced in the standard manner by applying an average of all the scores. This score was treated as an output.

We ran each model through the standard GLUE benchmark by training them on the \texttt{train} split of the dataset and evaluating them on the \texttt{eval} split; in doing so we generated several dozen metrics for each model. These metrics include standard metrics that capture model throughput, running time and performance; a condensed representative summary is presented in Table~\ref{tab:metrics}.

In practical applications of DEA, if the analysis uses far more inputs and outputs than the number of DMUs, then the typical outcome classifies all DMUs as Pareto efficient. There can still be value in analyses where this happens, this is atypical and we wish to avoid it.  A rule of thumb advises that the number of DMUs should be at least twice the number of inputs and outputs considered~\cite{COOK20141,GOLANY1989237}.  Following this advice, we run an analysis with just two inputs and two outputs. The inputs we use are 
$\log{\left(\textrm{\# trainable params}\right)}$ and total train runtime. The outputs were average score across all GLUE tasks and
average eval throughput (samples/second).

We nonlinearly transform the number of trainable parameters by applying $\log$ to it for two reasons. First, there is a large disparity between the number of trainable parameters that the simple models have, and the number of trainable parameters that the transformer models have. The result of this gap is that the feature effectively becomes a binary indicator of whether the model is a transformer or not, and this is not what we would like the feature to convey. The second reason is based on empirical observations about performance. Informally, we expect performance to be a sublinear function of model size. That is, model performance should improve as a function of model size, but with decreasing returns.

We ran our experiments via Google Cloud Platform's Vertex AI Pipelines.  Transformer models were trained on \texttt{n1-highmem-8} instances (8 vCPUs, 52 GiB memory) and one NVIDIA T4 GPU with CUDA toolkit version 11.2. Non-transformer models were trained on \texttt{e2-standard-4} instances (4 vCPUs, 16 GiB memory).  All experiments used Python 3.8 and, at the time of writing, the latest versions of major libraries\footnote{ The libraries and their versions are: (\texttt{torch}, {1.11.0}), (\texttt{transformers}, {4.20.1}), and (\texttt{scikit-learn},  {1.1.1}).}.  Our experiment script was a modified version of the \texttt{run\_glue.py} script included with Hugging Face's examples \footnote{\url{https://github.com/huggingface/transformers/blob/v4.20.1/examples/pytorch/text-classification/run_glue.py}}.
Runtimes for all tasks varied from minutes to hours depending on the task and model but all experiments were completed within 24 hours.

After generating the model metrics, we constructed the relevant linear programs described in Section~\ref{sec:math-setup}, and we solved them using Gurobi version 9.5.2.

The results shown in Table~\ref{tab:results} use the following definitions.
The column headed ``CCR score'' reports the optimal objective value  of the program in Eq.~\eqref{CCR envelopment}, and the ``BCC score'' reports the optimal objective value of the program in Eq.~\eqref{BCC envelopment}.  The column headed ``scale efficiency'' reports the ratio of the two optimal values, and is defined explicitly in Eq.~\eqref{Scale efficiency}. The column ``CCR eff.''~indicates whether the  optimal solution to \eqref{CCR envelopment} has zero slack, and reports the result of solving Eq.~\eqref{CCR envelopment slacks}. The column headed ``BCC eff.'' indicates whether the optimal solution to Eq.~\eqref{BCC envelopment} has zero slack and requires solving Eq.~\eqref{BCC envelopment slacks} to make the determination. Note that several models exhibit a BCC score of 1.000 while not being BCC-efficient, i.e., they are weakly efficient. Finally, the ``ret.\ to scale'' column containing either CRS or DRS (IRS does not occur here) reports the results of Eq.~\eqref{BCC multiplier} and Eq.~\eqref{BCC returns to scale}. CRS indicates constant returns to scale, which corresponds to item~\ref{item:crs} in the list appearing just prior to Eq.~\eqref{BCC returns to scale}. DRS indicates decreasing returns to scale. In this case, DRS corresponds to item~\ref{item:drs} in that same list.
% pbpaste | sed 's/stsb//' | sed 's/_//' | sed 's/_/\\_/g'

As a result, of this table, it is clear that the models \texttt{glove-50-linear}, \texttt{tfidf-1000-linear}, and \texttt{roberta-base} with \texttt{lr=1e-4} perform well overall. It is also clear that the BCC equations provide a view of model performance that benefits the more complex models. This confirms the general intuition that large changes in model size, complexity and other inputs yield incremental improvements in performance. Additionally, it shows that \texttt{bert-large-uncased} models are suboptimal, requiring a lot of time and space in exchange for performance that is similar to that of other models.
\begin{table*}[ht!]
    \centering
    \small
    \begin{tabular}{lrrr}
    \toprule
    Metric & Percentiles: 25 & 50 & 75\\
    \midrule
eval\_steps\_per\_second      & 0.258            & 0.295         & 0.561  \\ 
train\_samples\_per\_second   & 70.223           & 101.606       & 143.307 \\
num\_trainable\_params        & 82119169         & 109483009     & 124646401  \\
eval\_samples\_per\_second    & 129.03           & 147.476       & 280.338 \\
eval\_runtime                 & 5.3507           & 10.1712       & 11.6252 \\
train\_runtime                & 4011.6           & 5658.1        & 8186.7 \\
train\_steps\_per\_second     & 0.828            & 1.148         & 1.674 \\
eval\_combined\_score         & 0.867            & 0.880         & 0.894 \\
eval\_spearmanr               & 0.866            & 0.878         & 0.893 \\
eval\_pearson                 & 0.868            & 0.881         & 0.895 \\
% inv\_eval\_loss               & 1.765            & 1.889         & 2.175 \\
% inv\_train\_loss              & 1.020            & 1.729         & 3.240 \\
         \bottomrule
    \end{tabular}
    \caption{Representative data for the {stsb} tests. The five models tested are: bert-base-uncased,
bert-large-uncased,
distilbert-base-uncased,
distilroberta-base, and
roberta-base with three different learning rates $10^{-5}$, $10^{-4}$ and $10^{-3}$. In addition to stsb, the other tests are
mrpc,
qqp,
wnli,
rte,
mnli,
cola,
sst2, and
qnli. For each distinct model, each test, and each learning rate, similar metrics are generated, for a total of over 100 different metrics.}
    \label{tab:metrics}
\end{table*}

\begin{table*}[ht!]
\centering
\small
\begin{tabular}{lrrrcclr}
\toprule
{} &  $\theta_{CCR}^*$ &  $\theta_{BCC}^*$ &  $SE$ &  CCR eff. &  BCC eff. & RTS & GLUE score \\
\midrule
glove-50-linear                                  &   1.000 &   1.000 &          1.000 &              + &              + &              $\rightarrow$ & 0.408 \\
tfidf-1000-linear                                &   1.000 &   1.000 &          1.000 &              + &              + &              $\rightarrow$ & 0.591 \\
roberta-base, lr=1e-4%0.000100   
&   0.501 &   1.000 &          0.501 &                &              + &              $\downarrow$ & 0.830 \\ 
distilroberta-base, lr=1e-5%0.000010  
&   0.499 &   1.000 &          0.499 &                &              + &              $\downarrow$ & 0.815 \\ 
tfidf-10000-linear                               &   0.999 &   1.000 &          0.999 &                &                &              & 0.591 \\
tfidf-5000-linear                                &   0.999 &   1.000 &          0.999 &                &                &              & 0.591 \\
tfidf-500-linear                                 &   0.999 &   1.000 &          0.999 &                &                &              & 0.610 \\
tfidf-15000-linear                               &   0.999 &   1.000 &          0.999 &                &                &              & 0.591 \\
glove-100-linear                                 &   0.952 &   0.961 &          0.990 &                &                &              & 0.455 \\
roberta-base, lr=1e-5 %0.000010  
&   0.460 &   0.919 &          0.500 &                &                &              & 0.807 \\
bert-base-uncased, lr=1e-4%0.000100 
&   0.486 &   0.913 &          0.533 &                &                &              & 0.799 \\
distilroberta-base, lr=1e-4%0.000100
&   0.479 &   0.863 &          0.555 &                &                &              & 0.782 \\
glove-200-linear                                 &   0.841 &   0.845 &          0.996 &                &                &              & 0.446 \\
distilbert-base-uncased, lr=1e-5%0.000010 
&   0.460 &   0.842 &          0.546 &                &                &              & 0.579 \\
bert-base-uncased, lr=1e-5%0.000010 
&   0.437 &   0.841 &          0.519 &                &                &              & 0.789 \\
bert-large-uncased, lr=1e-5%0.000010  
&   0.385 &   0.835 &          0.461 &                &                &              & 0.800 \\
glove-300-linear                                 &   0.827 &   0.834 &          0.991 &                &                &           & 0.460 \\
distilbert-base-uncased, lr=1e-3%0.001000 
&   0.466 &   0.819 &          0.569 &                &                &              & 0.740 \\
distilbert-base-uncased, lr=1e-4%0.000100    
&   0.473 &   0.803 &          0.588 &                &                &              & 0.769 \\
bert-large-uncased, lr=1e-4%0.000100  
&   0.411 &   0.741 &          0.554 &                &                &              & 0.772 \\
distilroberta-base, lr=1e-3%0.001000    
&   0.452 &   0.679 &          0.665 &                &                &              & 0.729 \\
roberta-base, lr=1e-3%0.001000 
&   0.436 &   0.654 &          0.666 &                &                &              & 0.729 \\
bert-base-uncased, lr=1e-3%0.001000     
&   0.410 &   0.543 &          0.756 &                &                &              & 0.703 \\
bert-large-uncased, lr=1e-3%0.001000  
&   0.225 &   0.312 &          0.721 &                &                &              & 0.378 \\
\bottomrule
\end{tabular}
\caption{Efficiency scores, returns to scale characterizations of BCC-efficient models, and GLUE scores (average performance across tasks). Models are ranked first by their BCC score, then by their CCR score. Returns to scale characteristics (increasing = $\uparrow$, decreasing = $\downarrow$, constant = $\rightarrow$) indicated only for BCC-efficient models.}
\label{tab:results}
\end{table*}

\section{Conclusions and Future Work}
We have applied  Data Envelopment Analysis to the challenge of quantifying the trade-off that exists between model performance and resource demands. We base this analysis on standard high-dimensional summary statistics that describe each model. We apply DEA to the analysis of 14 natural language models, and from this analysis we identify both simple and transformer-based models that effectively balance the competing objectives.

We demonstrate that the method is feasible and scales well. Future work can refine the approach presented above in several directions. First, specifics of our analysis can be modified by selecting different sets of inputs and outputs, or by selecting different ways of normalizing the inputs and outputs. Although DEA is a quantitative framework, there is much subjectivity in how the analysis is set up and interpreted. Second, it would be interesting to consider a more principled approach to the normalization of inputs and output attributes used in the analysis. We take the $\log$ of the number of trainable parameters to amplify the difference between models where the number of parameters is small, as well as to capture diminishing resource cost once models are sufficiently large. For future work, one may apply $\exp$ to achieve the opposite effect. In addition, for attributes that take on negative values, since DEA assumes semipositive data, one may consider splitting the attribute into its positive and negative parts. Third, we have only considered input-oriented models, and so inherent in our approach is the goal of minimizing input consumption while maintaining best-in-class performance. An output-oriented approach is conversely interested in holding input resources constant while producing superior results. We leave investigation of these types of models to future work. Finally, it seems possible that DEA might be integrated into the training process, where the analysis is used to direct training time, parameter size, performance criteria.  Due to the high-dimensional nature of language model descriptions, we believe that DEA is  well-suited for language model~{assessment}.

\section*{Acknowledgments}
We thank the anonymous reviewers for their insightful comments and suggestions. This work was done while Zachary Zhou was an intern at American Family Insurance. American Family Insurance sponsored this work.

%\bibliography{main}

\begin{thebibliography}{36}
\expandafter\ifx\csname natexlab\endcsname\relax\def\natexlab#1{#1}\fi

\bibitem[{Ali(1990)}]{ALI1990157}
Agha~Iqbal Ali. 1990.
\newblock \href {https://doi.org/https://doi.org/10.1016/0198-9715(90)90020-T}
  {Data envelopment analysis: Computational issues}.
\newblock \emph{Computers, Environment and Urban Systems}, 14(2):157--165.

\bibitem[{Ali(1993)}]{ALI199361}
Agha~Iqbal Ali. 1993.
\newblock \href {https://doi.org/https://doi.org/10.1016/0377-2217(93)90008-B}
  {Streamlined computation for data envelopment analysis}.
\newblock \emph{European Journal of Operational Research}, 64(1):61--67.

\bibitem[{Banker et~al.(1984)Banker, Charnes, and Cooper}]{Banker1984}
R.~D. Banker, A.~Charnes, and W.~W. Cooper. 1984.
\newblock \href {https://doi.org/10.1287/mnsc.30.9.1078} {Some models for
  estimating technical and scale inefficiencies in data envelopment analysis}.
\newblock \emph{Management Science}, 30(9):1078--1092.

\bibitem[{Carter and Koopmans(1952)}]{Carter_1952}
C.~F. Carter and T.~C. Koopmans. 1952.
\newblock \href {https://doi.org/10.2307/2226909} {Activity analysis of
  production and allocation.}
\newblock \emph{The Economic Journal}, 62(247):625.

\bibitem[{Charnes et~al.(1995)Charnes, Cooper, Lewin, and
  Seiford}]{charnes1995data}
A.~Charnes, W.W. Cooper, A.Y. Lewin, and L.M. Seiford. 1995.
\newblock \href {https://books.google.com/books?id=K8zE0YnhREAC} {\emph{Data
  Envelopment Analysis: Theory, Methodology, and Applications}}.
\newblock Springer Netherlands.

\bibitem[{Charnes et~al.(1978)Charnes, Cooper, and
  Rhodes}]{charnes1978measuring}
Abraham Charnes, William~W Cooper, and Edwardo Rhodes. 1978.
\newblock Measuring the efficiency of decision making units.
\newblock \emph{European journal of operational research}, 2(6):429--444.

\bibitem[{Cook et~al.(2014)Cook, Tone, and Zhu}]{COOK20141}
Wade~D. Cook, Kaoru Tone, and Joe Zhu. 2014.
\newblock \href {https://doi.org/https://doi.org/10.1016/j.omega.2013.09.004}
  {Data envelopment analysis: Prior to choosing a model}.
\newblock \emph{Omega}, 44:1--4.

\bibitem[{Cooper et~al.(2004)Cooper, Seiford, and Zhu}]{handbook2004}
William~W. Cooper, Lawrence~M. Seiford, and Joe Zhu, editors. 2004.
\newblock \href {https://doi.org/10.1007/b105307} {\emph{Handbook on Data
  Envelopment Analysis}}.
\newblock Springer {US}.

\bibitem[{Cooper et~al.(2007)Cooper, Seiford, and Tone}]{cooper2007data}
W.W. Cooper, L.M. Seiford, and K.~Tone. 2007.
\newblock \href {https://books.google.com/books?id=XW0sswC0RzsC} {\emph{Data
  Envelopment Analysis: A Comprehensive Text with Models, Applications,
  References and DEA-Solver Software}}.
\newblock Springer US.

\bibitem[{Devlin et~al.(2019)Devlin, Chang, Lee, and
  Toutanova}]{DBLP:conf/naacl/DevlinCLT19}
Jacob Devlin, Ming{-}Wei Chang, Kenton Lee, and Kristina Toutanova. 2019.
\newblock \href {https://doi.org/10.18653/v1/n19-1423} {{BERT:} pre-training of
  deep bidirectional transformers for language understanding}.
\newblock In \emph{Proceedings of the 2019 Conference of the North American
  Chapter of the Association for Computational Linguistics: Human Language
  Technologies, {NAACL-HLT} 2019, Minneapolis, MN, USA, June 2-7, 2019, Volume
  1 (Long and Short Papers)}, pages 4171--4186. Association for Computational
  Linguistics.

\bibitem[{Doyle and Green(1994)}]{doyle1994efficiency}
John Doyle and Rodney Green. 1994.
\newblock Efficiency and cross-efficiency in dea: Derivations, meanings and
  uses.
\newblock \emph{Journal of the operational research society}, 45(5):567--578.

\bibitem[{Dror et~al.(2017)Dror, Baumer, Bogomolov, and
  Reichart}]{dror-replicability-analysis}
Rotem Dror, Gili Baumer, Marina Bogomolov, and Roi Reichart. 2017.
\newblock \href {https://transacl.org/index.php/tacl/article/view/1241}
  {Replicability analysis for natural language processing: Testing significance
  with multiple datasets}.
\newblock \emph{Transactions of the Association for Computational Linguistics},
  5(0):471--486.

\bibitem[{Dror et~al.(2019)Dror, Shlomov, and Reichart}]{dror-etal-2019-deep}
Rotem Dror, Segev Shlomov, and Roi Reichart. 2019.
\newblock \href {https://doi.org/10.18653/v1/P19-1266} {Deep dominance - how to
  properly compare deep neural models}.
\newblock In \emph{Proceedings of the 57th Annual Meeting of the Association
  for Computational Linguistics}, pages 2773--2785, Florence, Italy.
  Association for Computational Linguistics.

\bibitem[{Emrouznejad et~al.(2016)Emrouznejad, Banker, Ray, and
  Chen}]{dea2016aliraylei}
Ali Emrouznejad, Rajiv Banker, Subhash Ray, and Lei Chen. 2016.
\newblock \emph{Recent Applications of Data Envelopment Analysis}.
\newblock Proceedings of the 14th International Conference on Data Envelopment
  Analysis.

\bibitem[{Fowlkes and Mallows(1983)}]{fowlkesmallows}
E.~B. Fowlkes and C.~L. Mallows. 1983.
\newblock \href {https://doi.org/10.1080/01621459.1983.10478008} {A method for
  comparing two hierarchical clusterings}.
\newblock \emph{Journal of the American Statistical Association},
  78(383):553--569.

\bibitem[{Golany and Roll(1989)}]{GOLANY1989237}
B~Golany and Y~Roll. 1989.
\newblock \href {https://doi.org/https://doi.org/10.1016/0305-0483(89)90029-7}
  {An application procedure for dea}.
\newblock \emph{Omega}, 17(3):237--250.

\bibitem[{Guerrero et~al.(2022)Guerrero, Aparicio, and
  Valero-Carreras}]{dea4ml}
Nadia~M. Guerrero, Juan Aparicio, and Daniel Valero-Carreras. 2022.
\newblock \href {https://doi.org/10.3390/math10060909} {Combining data
  envelopment analysis and machine learning}.
\newblock \emph{Mathematics}, 10(6).

\bibitem[{{Hugging Face}(2022)}]{huggingface}
{Hugging Face}. 2022.
\newblock {Models -- Hugging Face}.
\newblock \url{https://huggingface.co/models}.
\newblock Accessed: 2022-08-05.

\bibitem[{Khezrimotlagh et~al.(2019)Khezrimotlagh, Zhu, Cook, and
  Toloo}]{KHEZRIMOTLAGH20191047}
Dariush Khezrimotlagh, Joe Zhu, Wade~D. Cook, and Mehdi Toloo. 2019.
\newblock \href {https://doi.org/https://doi.org/10.1016/j.ejor.2018.10.044}
  {Data envelopment analysis and big data}.
\newblock \emph{European Journal of Operational Research}, 274(3):1047--1054.

\bibitem[{Lin(2004)}]{rouge}
Chin-Yew Lin. 2004.
\newblock Rouge: A package for automatic evaluation of summaries.
\newblock \emph{Proceedings of the ACL Workshop: Text Summarization Braches Out
  2004}, page~10.

\bibitem[{Liu et~al.(2019)Liu, Ott, Goyal, Du, Joshi, Chen, Levy, Lewis,
  Zettlemoyer, and Stoyanov}]{DBLP:journals/corr/abs-1907-11692}
Yinhan Liu, Myle Ott, Naman Goyal, Jingfei Du, Mandar Joshi, Danqi Chen, Omer
  Levy, Mike Lewis, Luke Zettlemoyer, and Veselin Stoyanov. 2019.
\newblock \href {http://arxiv.org/abs/1907.11692} {Roberta: {A} robustly
  optimized {BERT} pretraining approach}.
\newblock \emph{CoRR}, abs/1907.11692.

\bibitem[{Matthews(1975)}]{MATTHEWS1975442}
B.W. Matthews. 1975.
\newblock \href {https://doi.org/https://doi.org/10.1016/0005-2795(75)90109-9}
  {Comparison of the predicted and observed secondary structure of t4 phage
  lysozyme}.
\newblock \emph{Biochimica et Biophysica Acta (BBA) - Protein Structure},
  405(2):442--451.

\bibitem[{Narang et~al.(2021)Narang, Chung, Tay, Fedus, Fevry, Matena, Malkan,
  Fiedel, Shazeer, Lan et~al.}]{narang2021transformer}
Sharan Narang, Hyung~Won Chung, Yi~Tay, William Fedus, Thibault Fevry, Michael
  Matena, Karishma Malkan, Noah Fiedel, Noam Shazeer, Zhenzhong Lan, et~al.
  2021.
\newblock Do transformer modifications transfer across implementations and
  applications?
\newblock In \emph{Conference on Empirical Methods in Natural Language
  Processing (EMNLP)}.

\bibitem[{Olesen and Petersen(2016)}]{OLESEN20162}
Ole~B. Olesen and Niels~Christian Petersen. 2016.
\newblock \href {https://doi.org/https://doi.org/10.1016/j.ejor.2015.07.058}
  {Stochastic data envelopment analysis—a review}.
\newblock \emph{European Journal of Operational Research}, 251(1):2--21.

\bibitem[{Papineni et~al.(2002)Papineni, Roukos, Ward, and
  Zhu}]{papineni2002bleu}
Kishore Papineni, Salim Roukos, Todd Ward, and Wei-Jing Zhu. 2002.
\newblock Bleu: a method for automatic evaluation of machine translation.
\newblock In \emph{Proceedings of the 40th annual meeting on association for
  computational linguistics}, pages 311--318. Association for Computational
  Linguistics.

\bibitem[{Pedregosa et~al.(2011)Pedregosa, Varoquaux, Gramfort, Michel,
  Thirion, Grisel, Blondel, Prettenhofer, Weiss, Dubourg, Vanderplas, Passos,
  Cournapeau, Brucher, Perrot, and Duchesnay}]{scikit-learn}
F.~Pedregosa, G.~Varoquaux, A.~Gramfort, V.~Michel, B.~Thirion, O.~Grisel,
  M.~Blondel, P.~Prettenhofer, R.~Weiss, V.~Dubourg, J.~Vanderplas, A.~Passos,
  D.~Cournapeau, M.~Brucher, M.~Perrot, and E.~Duchesnay. 2011.
\newblock Scikit-learn: Machine learning in {P}ython.
\newblock \emph{Journal of Machine Learning Research}, 12:2825--2830.

\bibitem[{Pennington et~al.(2014)Pennington, Socher, and
  Manning}]{pennington2014glove}
Jeffrey Pennington, Richard Socher, and Christopher~D. Manning. 2014.
\newblock \href {http://www.aclweb.org/anthology/D14-1162} {{GloVe: Global
  Vectors for Word Representation}}.
\newblock In \emph{Empirical Methods in Natural Language Processing (EMNLP)},
  pages 1532--1543.

\bibitem[{Phillips et~al.(1990)Phillips, Parsons, and Donoho}]{PHILLIPS1990167}
Fred Phillips, Ronald~G. Parsons, and Andrew Donoho. 1990.
\newblock \href {https://doi.org/https://doi.org/10.1016/0198-9715(90)90021-K}
  {Parallel microcomputing for data envelopment analysis}.
\newblock \emph{Computers, Environment and Urban Systems}, 14(2):167--170.

\bibitem[{Sanh et~al.(2019)Sanh, Debut, Chaumond, and
  Wolf}]{DBLP:journals/corr/abs-1910-01108}
Victor Sanh, Lysandre Debut, Julien Chaumond, and Thomas Wolf. 2019.
\newblock \href {http://arxiv.org/abs/1910.01108} {Distilbert, a distilled
  version of {BERT:} smaller, faster, cheaper and lighter}.
\newblock \emph{CoRR}, abs/1910.01108.

\bibitem[{Sun(2002)}]{SUN200251}
Shinn Sun. 2002.
\newblock \href {https://doi.org/https://doi.org/10.1016/S0038-0121(01)00010-6}
  {Measuring the relative efficiency of police precincts using data envelopment
  analysis}.
\newblock \emph{Socio-Economic Planning Sciences}, 36(1):51--71.

\bibitem[{Thanassoulis(1995)}]{THANASSOULIS1995641}
Emmanuel Thanassoulis. 1995.
\newblock \href {https://doi.org/https://doi.org/10.1016/0377-2217(95)00236-7}
  {Assessing police forces in england and wales using data envelopment
  analysis}.
\newblock \emph{European Journal of Operational Research}, 87(3):641--657.
\newblock Operational Research in Europe.

\bibitem[{Tsaples et~al.(2022)Tsaples, Papathanasiou, and Georgiou}]{dea4ml2}
Georgios Tsaples, Jason Papathanasiou, and Andreas~C. Georgiou. 2022.
\newblock \href {https://doi.org/10.3390/math10132277} {{An Exploratory DEA and
  Machine Learning Framework for the Evaluation and Analysis of Sustainability
  Composite Indicators in the EU}}.
\newblock \emph{Mathematics}, 10(13).

\bibitem[{Ulmer et~al.(2022)Ulmer, Bassignana, M{\"u}ller-Eberstein, Varab,
  Zhang, Hardmeier, and Plank}]{ulmer-experimental-standards}
{Dennis Thomas} Ulmer, Elisa Bassignana, Max M{\"u}ller-Eberstein, Daniel
  Varab, Mike Zhang, Christian Hardmeier, and Barbara Plank. 2022.
\newblock Experimental standards for deep learning research: A natural language
  processing perspective.

\bibitem[{Vaswani et~al.(2017)Vaswani, Shazeer, Parmar, Uszkoreit, Jones,
  Gomez, Kaiser, and Polosukhin}]{NIPS2017_3f5ee243}
Ashish Vaswani, Noam Shazeer, Niki Parmar, Jakob Uszkoreit, Llion Jones,
  Aidan~N Gomez, \L~ukasz Kaiser, and Illia Polosukhin. 2017.
\newblock \href
  {https://proceedings.neurips.cc/paper/2017/file/3f5ee243547dee91fbd053c1c4a845aa-Paper.pdf}
  {Attention is all you need}.
\newblock In \emph{Advances in Neural Information Processing Systems},
  volume~30. Curran Associates, Inc.

\bibitem[{Wang et~al.(2018)Wang, Singh, Michael, Hill, Levy, and
  Bowman}]{wang-etal-2018-glue}
Alex Wang, Amanpreet Singh, Julian Michael, Felix Hill, Omer Levy, and Samuel
  Bowman. 2018.
\newblock \href {https://doi.org/10.18653/v1/W18-5446} {{GLUE}: A multi-task
  benchmark and analysis platform for natural language understanding}.
\newblock In \emph{Proceedings of the 2018 {EMNLP} Workshop {B}lackbox{NLP}:
  Analyzing and Interpreting Neural Networks for {NLP}}, pages 353--355,
  Brussels, Belgium. Association for Computational Linguistics.

\bibitem[{Yule(1912)}]{yule}
G.~Udny Yule. 1912.
\newblock \href {http://www.jstor.org/stable/2340126} {On the methods of
  measuring association between two attributes}.
\newblock \emph{Journal of the Royal Statistical Society}, 75(6):579--652.

\end{thebibliography}

% NOTE: 
% CFP does allow supplementary appendices https://eval4nlp.github.io/2022/cfp.html
% \appendix
% \section{Appendix}
% \label{sec:appendix}

% \input{appendix}

\end{document}